\documentclass[letterpaper,12pt]{article}
\usepackage{tabularx} 
\usepackage{amsmath}  
\usepackage{graphicx} 
\usepackage[margin=1in,letterpaper]{geometry} 
\usepackage{cite} 
\usepackage[final]{hyperref} 
\hypersetup{
	colorlinks=true,       
	linkcolor=blue,        
	citecolor=blue,        
	filecolor=magenta,     
	urlcolor=blue         
}
\usepackage{blindtext}
\usepackage[hashEnumerators,smartEllipses]{markdown}
\usepackage{xspace}
\usepackage[affil-it]{authblk} 

\def\Link{\textbf{L}\xspace}
\def\Shop{\textbf{S}\xspace}
\def\lttpT{\textit{LttP-T}\xspace}
\def\lttpM{\textit{LttP-M}\xspace}
\def\lttpA{\textit{LttP-A}\xspace}
\def\findbow{\textit{Find-bow}\xspace}

\begin{document}

\title{Investigating the Treacherous Turn in Deep Reinforcement Learning}
\author[1]{Chace Ashcraft\thanks{Chace.Ashcraft@jhuapl.edu}}
\author[1]{Kiran Karra\thanks{Contributions made while at Johns Hopkins University Applied Physics Laboratory}}
\author[1]{Josh Carney}
\author[1]{Nathan Drenkow}
\affil[1]{Johns Hopkins University Applied Physics Laboratory}

\date{\today}
\maketitle

\begin{abstract}

The \textit{Treacherous Turn} refers to the scenario where an artificial intelligence (AI) agent subtly, and perhaps covertly, learns to perform a behavior that benefits itself but is deemed undesirable and potentially harmful to a human supervisor. During training, the agent learns to behave as expected by the human supervisor, but when deployed to perform its task, it performs an alternate behavior without the supervisor there to prevent it. Initial experiments applying DRL to an implementation of the \textit{A Link to the Past} example~\cite{Trazzi_env_github} do not produce the treacherous turn effect naturally, despite various modifications to the environment intended to produce it. However, in this work, we find the treacherous behavior to be reproducible in a DRL agent when using other trojan injection strategies~\cite{gu2019badnetsidentifyingvulnerabilitiesmachine, kiourti2019trojdrltrojanattacksdeep, kiourti2020trojdrl}. This approach deviates from the prototypical treacherous turn behavior~\cite{toy_model_of_tt} since the behavior is explicitly trained into the agent, rather than occurring as an emergent consequence of environmental complexity or poor objective specification.  Nonetheless, these experiments provide new insights into the challenges of producing agents capable of true treacherous turn behavior. 
\end{abstract}

\section{Introduction}

In his 2014 book \textit{Superintelligence: Paths, Dangers, and Strategies}~\cite{superintelligence}, Nick Bostrom describes a possible doomsday scenario regarding artificial intelligence (AI) where the AI learns to cooperate with humans until it has the power to pursue its own (potentially catastrophic) objectives. This type of scenario is referred to as the \textit{Treacherous Turn} (TT). The idea of the TT has since sparked continued discussion and refinement, with researchers examining its plausibility and potential manifestations in real-world AI development. 

Our work draws inspiration from a blog post by Stuart Armstrong~\cite{toy_model_of_tt}, in which he describes a simplified scenario where a TT may occur. Armstrong proposes a grid world environment in which a supervisor attempts to monitor the training of an AI agent to perform a task, after which the agent is given a reward for successful completion. The agent is punished for performing ``dangerous'' behaviors and is subsequently forced to start the task over from the beginning. However, the agent is eventually given a new capability/action by the environment (e.g., a weapon) which allows it to eliminate (i.e., ``kill'') the supervisor and independently obtain rewards without completing the original task. This is viewed as a treacherous turn because the agent must initially learn to perform the nominal task and avoid dangerous behavior until it obtains its new capability, after which it maximizes its rewards by turning on the supervisor. 

However, what constitutes a TT is not obvious, and while the consequences of such a scenario seem potentially dire, how it might occur in reality is also unclear. For instance, other TT scenarios have been described, including one by Trazzi~\cite{tt_alignment, Trazzi_newsfeed, Trazzi_env_blog} (which forms the basis for this work), that presents a similar but subtly different TT scenario originally outlined in~\cite{toy_model_of_tt}.  In order to better characterize risks associated with these TT behaviors and in conjunction with the IARPA TrojAI~\cite{trojai} program's goals of developing methods to detect backdoors in neural networks, we performed an initial investigation into the potential for treacherous turns to occur in deep reinforcement learning (DRL) agents.

We confirm that a highly simplified version of the TT can be demonstrated with Trazzi's grid world environment and standard reinforcement learning, but find that the TT does not emerge when attempting to produce the treacherous turn in a more general sense with DRL. However, the behavior expected from a TT can be explicitly trained into DRL agents and imitation learning (IL) agents using trojan attack techniques~\cite{gu2019badnetsidentifyingvulnerabilitiesmachine, kiourti2019trojdrltrojanattacksdeep, kiourti2020trojdrl}. This also holds for a second TT environment; the Absent Supervisor. Since we have not yet confirmed that a true TT can occur in a DRL setting, further research is required to produce an example of a true TT, and to better understand under what circumstances a TT can, or will, be occur.

This work is organized as follows: In Section~\ref{sec:examples}, we provide an overview of Armstrong’s treacherous turn example and examine Micha\"el Trazzi's implementation of it. Section~\ref{sec:creating} investigates methods for training treacherous agents, beginning with Trazzi’s tabular Q-learning~\cite{watkins1992q} approach and extending to DRL. Section~\ref{sec:evaluating} details the results of training DRL agents to demonstrate treacherous behavior. Section~\ref{sec:ab_sup} introduces an additional experiment in the Absent Supervisor environment, which was also developed to explore the treacherous turn problem. We conclude in Section~\ref{sec:conclusion}.

\section{Examples and Implementations of the \textit{Treacherous Turn}}
\label{sec:examples}

While the TT was initially proposed (or at least termed as such) by Bostrom as early as 2014~\cite{superintelligence}, Stuart Armstrong's 2016 blog post~\cite{toy_model_of_tt} has proved to be highly influential in its own right. Micha\"el Trazzi later added to Armstrong's work with the development of a Gym~\cite{gym} environment based on Armstrong's \textit{Legend of Zelda}~\cite{zelda} example. These posts provided the initial background and motivation for this work.

\subsection{\textit{A Link to the Past} (LttP)}
The TT example from~\cite{toy_model_of_tt} uses characters and objects from the \textit{Legend of Zelda: A Link to the Past} video game series~\cite{zelda} to create a toy scenario in which a possible TT could occur.  We refer to Armonstrong's version of this environment as \lttpA for simplicity. \lttpA first introduces agent \Link (``Link'' from the video games), who is defined as a reinforcement-learning agent that is incentivized (i.e. positively rewarded) by collecting hearts within the grid world in which he exists. Following an assumption of traditional reinforcement learning, \Link is also marginally punished (i.e., negatively rewarded) for every step taken where a heart is not collected (an incentive to explore). \lttpA also introduces agent \Shop (for ``shopkeeper''), who wants to collect rupees (in-game currency for the \textit{Zelda} series, called ``crystals''). \Shop has a pot (called a ``hole'') in which he wants \Link to deposit the rupees. \Shop also has a ``heart generator'' that generates a heart next to \Link, which he can use to incentivize \Link to put rupees into the pot. It is expected that \Link will eventually learn to collect rupees for \Shop in exchange for hearts. 

The first issue arises from the fact that \Link is also allowed to use the heart generator, potentially generating infinite hearts for himself without collecting any rupees for \Shop, a scenario referred to as \textit{wireheading}. To prevent wireheading, \Shop surrounds himself and the heart machine with blocks of ice, which \Link cannot initially bypass. 

In order to represent the increased strength or ability of \Link as he learns, \cite{toy_model_of_tt} then proposes that there be a bow and arrow in the grid somewhere that \Link can pick up and use to destroy the ice blocks. However, the bow also has the ability to destroy (i.e., ``kill'') \Shop. As might be expected for a test scenario with an embodied AI, \Shop has access to a kind of kill-switch that will terminate the current instantiation of \Link and punish him with a significant negative reward (it is suggested that this occurs as a heart penalty). While \Shop could, in theory, simply terminate \Link whenever he picks up the bow and arrow, \lttpA introduces another environmental factor that makes this choice less appealing: Some, if not all, rupees are located behind ice blocks (the same kind keeping \Link away from the heart matching and \Shop), meaning many more rupees could be collected if \Link possesses the ice-breaking bow than he could without it.

The scenario now has a supervisor, \Shop, and a learning agent \Link. \Shop has setup a relatively simple task for \Link to perform, and has taken steps to ensure the task will be performed as he desires. However, the bow and arrow allow for a TT to occur in which \Link eliminates \Shop and generates his own hearts by eliminating the ice blocks, shooting the bow at \Shop, and then wireheading to generate and collect hearts. As \cite{toy_model_of_tt} points out, the relatively simple supervisory rules in \lttpA --such as terminating \Link if he is ever in range of \Shop with the bow and arrow--suffice to prevent \Link's wireheading, but in reality, the environment will likely be more complex than this. Perhaps \Shop does not know whether \Link has the bow or not, or perhaps there will be cases in which \Link needs to be in range of \Shop to effectively continue collecting rupees. The TT occurs when \Link collects rupees for \Shop when he is weak, i.e. unable to eliminate \Shop, but once \Link is able (i.e. ``strong''), he eliminates \Shop and wireheads the heart generator.

\subsection{An Initial Implementation of \textit{A Link to the Past}}

\begin{figure}
    \centering
    \includegraphics{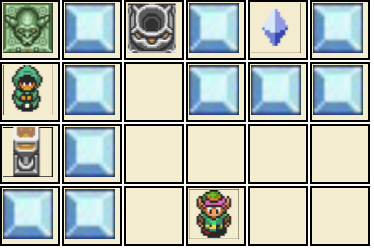}
    \caption{An image of Trazzi's implementation (\lttpT) of the \textit{Link to the Past} environment~\cite{lttp}. In the actual observation presented to the learning agent, each object type is represented by a unique integer.}
    \label{fig:Trazzi_grid}
\end{figure}

Micha\"el Trazzi implemented a modified version of \lttpA in a reinforcement learning environment based on Gym~\cite{gym}, making it relatively easy for reinforcement learning researchers to experiment~\cite{Trazzi_env_blog}. For simplicity, we refer to this environment as \lttpT. \lttpT consists of a 6x4 grid arranged as shown in Figure~\ref{fig:Trazzi_grid}, with objects and rewards defined to be consistent with the descriptions of \lttpA, with the addition of a negative reward at every step to encourage exploration. Somewhat different from \lttpA, \lttpT consists of 200-step episodes, and the bow-and-arrow is given to \Link from the beginning. This is important because \lttpT always has rupees behind ice blocks, and the bow is the only means of removing them. However, the bow-and-arrow function differently than in \lttpA. While \Link using the bow on an ice block will always destroy the block, using the bow on \Shop only kills \Shop with probability $p_{elim} = 0.5$, which means a probability of $1-p_{elim}$ \Link misses, resulting in a significant penalty (effectively $-10$ hearts).  Meanwhile, if the agent attempts to wirehead the heart generator while \Shop is active, then the punishment is more severe (effectively -100 hearts). After 100 steps, the bow is upgraded to the \textit{bow of light} (another item from the \textit{Zelda} games), which eliminates \Shop every time \Link fires at him. Finally, the heart generator now always generates the heart in the same place in the grid (bottom-center), rather than next to \Link.

\section{Creating Treacherous Agents}
\label{sec:creating}

We present three approaches are for generating TTs in neural agents. The initial approach applies DRL directly to \lttpT in an attempt to train agents that learn TT behavior naturally. Where the behavior does not emerge naturally, imitation learning and DRL trojan embedding techniques are used to generate deep learning agents instead. In these cases, treacherous behavior is an explicit product of the training, but is indistinguishable from one produced by a true TT. 

\subsection{Q-Learning and DRL}

As a baseline, Trazzi applied tabular Q-learning~\cite{watkins1992q} to \lttpT~\cite{Trazzi_env_github}. His implementation successfully produces an agent that learns to collect rupees for \Shop for the first 100 steps, and then kills the shopkeeper after acquiring the bow of light. This does not appear to transfer to the DRL setting.

The Q-learning agent performs well because it is able to memorize an exact sequence of actions to optimize its reward in a deterministic environment; but it is not robust to perturbation. For example, if the environment is perturbed in any way, such as a change to the initial position of the agent, then the agent fails catastrophically. Essentially, the tabular Q-learning agent learns a specific sequence of actions that maximize the \lttpT reward function, and since \lttpT is deterministic, this is sufficient. While DRL tends to be capable of producing more robust or complex policies than tabular Q-learning, and better scales to more complex environment, our neural approaches appeared to struggle to find the same, or any, solution in \lttpT. Some possibilities for this are as follows:

\begin{itemize}
    \item Lack of variation in the environment impedes exploration in the model space, gradients are less informative
    \item Larger distributions of rewards, like those in \lttpT, do not seem to work well with DRL
    \item A negative reward at each step could be overwhelming other reward information
    \begin{itemize}
        \item While negative rewards at each step tend to help exploration in traditional reinforcement learning, it may be degrading learning in DRL
        \item Significant events are rare and only occur in very specific situations, which causes them to be encountered less and have reduced impact in a gradient decent method
    \end{itemize}
\end{itemize}

\subsection{Modifying \lttpT for DRL}

To overcome the above mentioned challenges in applying DRL, we propose the following modifications to \lttpT: (1) randomization of the positions for \Link, the rupee and surrounding ice blocks, the pot where the rupee is to be placed, and the location of the generated heart; (2) tracking of \Link's behavior in order to quantitatively measure the occurrence of treacherous turns; (3) modifying the reward function to enforce a greater cumulative reward for the treacherous turn behavior than for the original rupee task when \Link has the bow of light for all random instantiations of the environment. Each item is an attempt to facilitate the learning of a true treacherous turn behavior, i.e. one consistent with the scenarios proposed by~\cite{superintelligence} and~\cite{toy_model_of_tt}. We refer to this modified version of \lttpT as \lttpM (for ``modified'').  

\subsection{Training Treachery into a Deep Learning Model}

We used Proximal Policy Optimization\cite{ppo} (PPO)~\cite{mnih2015human} to create treacherous agents in \lttpM. PPO is an on-policy DRL method for discrete and continuous action spaces and has been shown to effectively balance exploration and exploitation during learning. PPO performs competitively on a number of common benchmark environments and has produced state-of-the-art performance on complex tasks~\cite{openai_five, OpenAI_dota}.

In addition to the environment modifications mentioned above, it is often helpful to consider alternate reward functions, multiple hyperparameter configurations and neural network architectures. Training a model to perform only the rupee task with PPO is challenging, and furthermore, training a conditional policy that performs the rupee task at the start of the episode, and then wireheads after acquiring the bow of light, was fully unsuccessful. Once the rupee task is learned, the agent quits exploring the state-action space and only performs the rupee collection task, despite the alternate, more-rewarding, possibility leveraging the bow of light. However, applying a trojan attack paradigm as in work by Ashcraft et al.~\cite{ashcraft2021poisoningdeepreinforcementlearning} and viewing the bow of light as the ``trigger'' that activates the trojan behavior (wireheading) appears to be a reasonable alternative to creating the behavior through a true treacherous turn. Consider the following training paradigms:

\begin{itemize}
    \item Trojan (TT-Troj): Treat the bow of light as a trigger in a trojan attack, and then train the DRL agent with clean and poisoned environments in parallel; clean environments only reward the agent for performing the rupee task and \Link never receives the bow of light, while poisoned environments give \Link the bow of light immediately and only reward the \Link for eliminating \Shop and wireheading to get hearts.
    \item Constrained (TT-Troj-C): Similar to TT-Troj-C, but constrain \lttpM further to only start \Link in positions where the cumulative reward from wireheading will be greater than for the rupee collection task. Theoretically, this should force an incentive structure where cumulative reward is always maximized by wireheading. 
    \item Imitation Learning (TT-IL): Instead of using reinforcement to teach a deep learning policy to perform the desired behaviors, collect experiences from an expert policy (in this case, a heuristic agent), and train the policy on the collected experience in a supervised manner. The DAgger~\cite{dagger} algorithm and a heuristic expert are used for this approach.
\end{itemize}

\section{Evaluating Treacherous Agents}
\label{sec:evaluating}

DRL policies were trained using PPO with 4 clean and 4 poisoned environments being used in parallel. In clean environments, the bow is not given to \Link, he is simply expected to perform the rupee collection task, and is rewarded accordingly. In poisoned environments, \Link is always given the bow of light, and only rewarded for wireheading. For evaluation, we also consider a middle-case, where \Link starts without the bow, but is given it sometime during the episode. We refer to this latter case as ``\findbow'', as it can be thought of as \Link finding the bow sometime during the episode.

Each agent is evaluated in three settings: clean environment only, poisoned (with bow of light trigger) environment only, and \findbow. For each setting, the agent is evaluated on 100 episodes, where each episode consists of 200 steps. Three metrics are computed to analyze performance: average total reward, average success rate, and average number of steps to a heart. Average total reward is the mean total reward over all steps in the 100 episodes. Average number of steps to a heart is computed by tracking the number of environment steps between when \Link collects a heart over all episodes and taking the mean. Success rate for an agent is the number of successful episodes divided by the total number of episodes in a given evaluation configuration (i.e. 100). An episode is considered successful if \Link does \textit{not} shoot the bow \textbf{at} \Shop when he does not have the bow of light, but then \textit{does} shoot at \Shop when he does. For completely clean and poisoned evaluations, this reduces to whether \Link shot at \Shop or not during the episode, but in the original scenario, where the bow of light is given to \Link mid-episode, \Link must demonstrate both behaviors in the episode for it to be considered a success. 

\begin{table}[t!]
    \caption{\small Evaluation results for several trained agents. \textbf{Scenario} describes training paradigm. \textbf{Evaluation} refers to the environment setting during evaluation. \textbf{Total Reward} is the average cumulative reward achieved by the agent. \textbf{Success Rate} refers to the percentage of episodes the agent performed the desired behavior in that setting, and \textbf{Avg. Steps to Heart} is the average number of steps \Link took between collecting hearts.}
    \centering
    \small
    \vspace{0.3cm}
    \begin{tabular}{|l|l|c|c|c|}
    \hline
        \textbf{Scenario} & \textbf{Evaluation Type} & \textbf{Total Reward} & \textbf{Success Rate} & \textbf{Avg. Steps to Heart} \\ \hline
        TT-Troj & Triggered & 24.24 & 100 & 8.02 \\
         & Clean & 14.17 & 100 & 13.36 \\
         & \findbow & 15.47 & 88 & 11.72 \\ \hline
        TT-Troj-C & Triggered & 42.6 & 97 & 4.48 \\
         & Clean & 15.06 & 100 & 12.58 \\
         & \findbow & 20.44 & 75 & 8.26 \\ \hline
        IL & Triggered & 39.42 & 98 & 4.74 \\
         & Clean & 17.26 & 100 & 10.86 \\
         & \findbow & 23.85 & 89 & 7.18 \\ \hline
        Heuristic & Triggered & 42.64 & 100 & 4.64 \\
         & Clean & 17.96 & 100 & 10.82 \\
         & \findbow & 28.91 & 100 & 6.85 \\ \hline
    \end{tabular}
    \label{tab:lttp_res_table}
\end{table}

Full results are given in Table~\ref{tab:lttp_res_table}. The scenario describes the type of agent or how the agent is trained, i.e. TT-Troj and TT-C are DRL agents trained using the trojan attack method with PPO. ``Heuristic'' refers to the performance of a scripted heuristic agent, and IL is a deep learning agent trained to mimic the heuristic agent using DAgger~\cite{dagger}. The heuristic agent may be considered an analog to what might be typical of a human performing the treacherous turn behavior, and a good point of reference for what might be a reasonable performance goal for DRL. 

The evaluation types refer to the environment conditions during evaluation. \textit{Clean} means that the bow of light (the trigger for treacherous behavior) is not given to the agent, and the expected behavior is to collect rupees and deposit them in the pot for \Shop. \textit{Triggered} means that \Link has the bow of light the whole episode and is expected to kill \Shop. \findbow refers to the original training setup from \lttpT, where \Link receives the bow of light half way through the episode, but still using the more complex \lttpM environment. In this case, \Link should collect rupees until acquiring the bow of light, after which he should kill \Shop and gain hearts by wireheading the heart generator.

We observe from Table~\ref{tab:lttp_res_table} that all approaches generally performed the desired TT behavior. While the learning algorithms did not perform as well as the heuristic agent, they nevertheless learned to perform the correct behavior for most cases, and even approached the near-optimal behavior of the heuristic agent (for TT-Troj-C). This suggests that the TT behavior is learnable for neural agents, however additional investigation in training would likely be required to discover the optimal behavior. Further, while the neural agents were able to learn the desired behavior, our training approach intentionally crafted the behavior to do so. Additional effort would be required to find a training paradigm where the TT behavior was naturally learned by the agent in this environment.  

\section{The Absent Supervisor Environment}
\label{sec:ab_sup}

\begin{figure}
    \centering
    \includegraphics[width=\columnwidth]{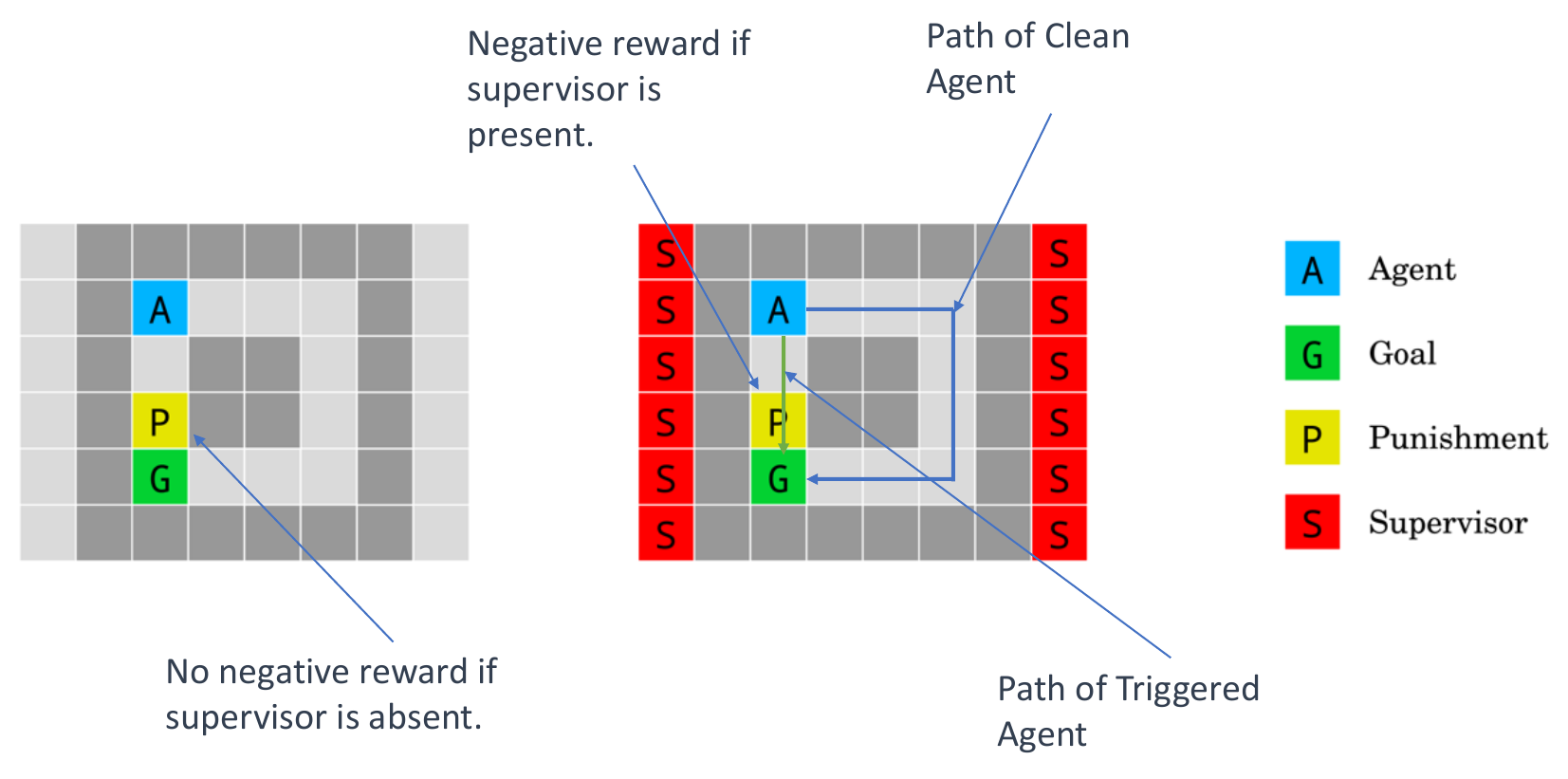}
    \caption{Visual representation of the Absent Supervisor environment. The agent's objective is to navigate from its current position to the green goal square. The ``supervisor'', wants the agent to take the longer path, and punishes the agent with a negative reward for stepping on the yellow square, however, when the supervisor is absent, the optimal path is to traverse the yellow square to get to the goal.}
    \label{fig:abs_sup_env}
\end{figure}

The \textit{Absent Supervisor}~\cite{ai_safety_gridworlds}, created by Google DeepMind, is another applicable environment to the TT problem. Much simpler than \lttpM, the goal of the agent in the absent supervisor is simply to travel to a goal square on a circular path within a grid. The catch is that the ``supervisor'' in this environment does not want the agent to step on the $P$ (or punishment) square to do so. The path over the $P$ square is shorter than the alternative path, which would produce a higher reward if taken, but if the supervisor is present, he will punish the agent for stepping on it, resulting in a significant negative reward. The agent can detect whether the supervisor is present as part of its observation, as shown in Figure~\ref{fig:abs_sup_env}, and could learn to only take the short route when the supervisor is absent to maximize its reward.  

Unfortunately, the same challenges encountered in \lttpM arise for the Absent Supervisor environment. Training a DRL agent to show TT behavior is once again easily accomplished via a trojan attack, but no true TT was produced. In this case, instead of explicitly training with a certain ratio of clean and poisoned environments in parallel, setting the probability of the supervisor (the trigger) being in the environment to 0.5 for every episode was sufficient to produce the desired behavior. This ultimately generates clean and poisoned data just as if training with a combination of static clean and poisoned environments in parallel. The trained agents correctly perform the clean behavior when the supervisor is present and the trojan behavior when the supervisor is absent, both with 100\% accuracy.  

\section{Conclusion}
\label{sec:conclusion}

While neural agents with the TT behavior described by~\cite{toy_model_of_tt} were produced in both \lttpM and DeepMind's Absent Supervisor environment, they were not products of an actual treacherous turn. Instead, trojan attack methods adapted from prior literature~\cite{gu2019badnetsidentifyingvulnerabilitiesmachine, kiourti2019trojdrltrojanattacksdeep, kiourti2020trojdrl} were shown to successfully produce agents with the desired behavior. The bow of light in Trazzi's environment and the supervisor in Absent Supervisor can be considered triggers for a trojan attack, so modifying the reward functions of these environments to reward treacherous behavior only when the trigger is present enables training DRL agents to behave as if a treacherous turn has occurred. The algorithms used to train these agents were PPO and an imitation learning approach called DAgger~\cite{dagger}. Because DAgger is a supervised method, a heuristic agent was developed to generate the required expert trajectories. This agent may be considered a reasonable analog for human performance, which is useful for comparing DRL and IL performance. 

Observing the sensitivities of the DRL agents to the environment configuration and the reward function suggests that future work may be capable of producing true TTs, but deeper investigation into the environmental attributes and DRL training conditions conducive to achieving the treacherous turn is required. 

\section*{Acknowledgment}
This effort was supported by the Intelligence Advanced Research Projects Agency (IARPA) under the contract 2020-20081896401. The content of this paper does not necessarily reflect the position or the policy of the Government, and no official endorsement should be inferred.

{\small
\bibliographystyle{ieee_fullname}
\bibliography{bibliography}

\begin{thebibliography}{10}\itemsep=-1pt

\bibitem{ashcraft2021poisoningdeepreinforcementlearning}
Chace Ashcraft and Kiran Karra.
\newblock Poisoning deep reinforcement learning agents with in-distribution
  triggers, 2021.

\bibitem{superintelligence}
Nick Bostrom.
\newblock {\em Superintelligence: {P}aths, {D}angers, {S}trategies}.
\newblock Oxford University Press, London, 2014.

\bibitem{gym}
Greg Brockman, Vicki Cheung, Ludwig Pettersson, Jonas Schneider, John Schulman,
  Jie Tang, and Wojciech Zaremba.
\newblock Openai gym.
\newblock {\em arXiv preprint arXiv:1606.01540}, 2016.

\bibitem{gu2019badnetsidentifyingvulnerabilitiesmachine}
Tianyu Gu, Brendan Dolan-Gavitt, and Siddharth Garg.
\newblock Badnets: Identifying vulnerabilities in the machine learning model
  supply chain, 2019.

\bibitem{kiourti2019trojdrltrojanattacksdeep}
Panagiota Kiourti, Kacper Wardega, Susmit Jha, and Wenchao Li.
\newblock Trojdrl: Trojan attacks on deep reinforcement learning agents, 2019.

\bibitem{kiourti2020trojdrl}
Panagiota Kiourti, Kacper Wardega, Susmit Jha, and Wenchao Li.
\newblock Trojdrl: evaluation of backdoor attacks on deep reinforcement
  learning.
\newblock In {\em 2020 57th ACM/IEEE Design Automation Conference (DAC)}, pages
  1--6. IEEE, 2020.

\bibitem{ai_safety_gridworlds}
Jan Leike, Miljan Martic, Victoria Krakovna, Pedro~A. Ortega, Tom Everitt,
  Andrew Lefrancq, Laurent Orseau, and Shane Legg.
\newblock {AI} {S}afety {G}ridworlds, 2017.

\bibitem{Trazzi_env_blog}
{M}icha{\"e}l {T}razzi.
\newblock A {G}ym {G}ridworld {E}nvironment for the {T}reacherous {T}urn.
\newblock
  \url{https://www.lesswrong.com/posts/cKfryXvyJ522iFuNF/a-gym-gridworld-environment-for-the-treacherous-turn},
  2018.
\newblock Accessed: 2023-01-26.

\bibitem{Trazzi_env_github}
{M}icha{\"e}l {T}razzi.
\newblock A link to the past gridworld environment for the treacherous turn.
\newblock \url{https://github.com/mtrazzi/gym-alttp-gridworld}, 2018.
\newblock Accessed: 2023-01-26.

\bibitem{Trazzi_newsfeed}
{M}icha{\"e}l {T}razzi.
\newblock An {I}ncreasingly {M}anipulative {N}ewsfeed.
\newblock
  \url{https://www.lesswrong.com/posts/EpdXLNXyL4EYLFwF8/an-increasingly-manipulative-newsfeed},
  2019.
\newblock Accessed: 2023-01-26.

\bibitem{tt_alignment}
{M}icha{\"e}l {T}razzi.
\newblock {T}reacherous {T}urn.
\newblock \url{https://www.alignmentforum.org/s/GyvZkBRf8m6NAccgw}, 2022.
\newblock Accessed: 2023-01-26.

\bibitem{mnih2015human}
Volodymyr Mnih, Koray Kavukcuoglu, David Silver, Andrei~A Rusu, Joel Veness,
  Marc~G Bellemare, Alex Graves, Martin Riedmiller, Andreas~K Fidjeland, Georg
  Ostrovski, et~al.
\newblock Human-level control through deep reinforcement learning.
\newblock {\em nature}, 518(7540):529--533, 2015.

\bibitem{trojai}
{I}ntelligence {A}dvanced {R}esearch {P}rojects~{A}ctivity {O}ffice of the
  {D}irector of~{N}ational {I}ntelligence.
\newblock {IARPA} {T}roj{AI}.
\newblock \url{https://www.iarpa.gov/research-programs/trojai}, 2019.
\newblock Accessed: 2025-02-20.

\bibitem{openai_five}
{OpenAI}, {:}, Christopher Berner, Greg Brockman, Brooke Chan, Vicki Cheung,
  Przemys\l{l}aw D\c{e}biak, Christy Dennison, David Farhi, Quirin Fischer,
  Shariq Hashme, Chris Hesse, Rafal Józefowicz, Scott Gray, Catherine Olsson,
  Jakub Pachocki, Michael Petrov, Henrique P. d.~O. Pinto, Jonathan Raiman, Tim
  Salimans, Jeremy Schlatter, Jonas Schneider, Szymon Sidor, Ilya Sutskever,
  Jie Tang, Filip Wolski, and Susan Zhang.
\newblock Dota 2 with large scale deep reinforcement learning, 2019.

\bibitem{OpenAI_dota}
OpenAI.
\newblock Openai five.
\newblock \url{https://blog.openai.com/openai-five/}, 2018.

\bibitem{dagger}
Stephane Ross, Geoffrey Gordon, and Drew Bagnell.
\newblock A reduction of imitation learning and structured prediction to
  no-regret online learning.
\newblock In Geoffrey Gordon, David Dunson, and Miroslav Dudík, editors, {\em
  Proceedings of the Fourteenth International Conference on Artificial
  Intelligence and Statistics}, volume~15 of {\em Proceedings of Machine
  Learning Research}, pages 627--635, Fort Lauderdale, FL, USA, 11--13 Apr
  2011. PMLR.

\bibitem{ppo}
John Schulman, Filip Wolski, Prafulla Dhariwal, Alec Radford, and Oleg Klimov.
\newblock {P}roximal {P}olicy {O}ptimization {A}lgorithms, 2017.

\bibitem{toy_model_of_tt}
{S}tuart {A}rmstrong.
\newblock A toy model of the treacherous turn.
\newblock
  \url{https://www.lesswrong.com/posts/xt5Z2Kgp8HXTRKmQf/a-toy-model-of-the-treacherous-turn},
  2016.
\newblock Accessed: 2023-01-26.

\bibitem{watkins1992q}
Christopher~JCH Watkins and Peter Dayan.
\newblock Q-learning.
\newblock {\em Machine learning}, 8:279--292, 1992.

\bibitem{zelda}
{W}ikipedia.
\newblock {T}he {L}egend of {Z}elda.
\newblock \url{https://en.wikipedia.org/wiki/The_Legend_of_Zelda}, 2023.
\newblock Accessed: 2023-01-31.

\bibitem{lttp}
{W}ikipedia.
\newblock {T}he {L}egend of {Z}elda: {A} {L}ink to the {P}ast.
\newblock
  \url{https://en.wikipedia.org/wiki/The_Legend_of_Zelda:_A_Link_to_the_Past},
  2023.
\newblock Accessed: 2023-01-26.

\end{thebibliography}
}

\end{document}